\documentclass[conference]{IEEEtran}
\IEEEoverridecommandlockouts
\usepackage{soul}
\usepackage{cite}
\usepackage{amsmath,amssymb,amsfonts}
\usepackage{algorithmic}
\usepackage{graphicx}
\usepackage{textcomp}
\usepackage{xcolor}
\usepackage{subcaption}
\usepackage{hyperref}
\usepackage{multirow}
\usepackage{breakurl}
\usepackage{booktabs}
\usepackage{balance}

\def\BibTeX{{\rm B\kern-.05em{\sc i\kern-.025em b}\kern-.08em
    T\kern-.1667em\lower.7ex\hbox{E}\kern-.125emX}}
\begin{document}

\newcommand{\edit}[1]{\textcolor{blue}{#1}}

\newcommand{\cm}[1]{\textcolor{red}{CM: #1}}

\newcommand{\daniel}[0]{\textcolor{red}{@Daniel }}

% \title{VOSA: Partially Observable Shared Autonomy via End-Effector Vision for Zero-Shot Manipulation Intent Recognition
% }

\title{Toward Zero-Shot User Intent Recognition in Shared Autonomy
}

% \title{Toward Zero-Shot Vision-Only Shared Autonomy}

%How far can off-the-shelf, zero-shot intent recognition get you in shared control?

%Investigating a Simple Approach to Zero-Shot Recognition of Possible Intents During Shared Autonomy

%Investigating a Simple Approach to Zero-Shot Manipulation Intent Recognition for Shared Autonomy

% Giving shared control robots eyes to see. 

% Moving beyond hard-coded intent sets in shared autonomy user studies. 

% A simple way to level up shared autonomy user studies via zero-shot intent set recognition.

\author{
\IEEEauthorblockN{Atharv Belsare$^*$}
\IEEEauthorblockA{\textit{University of Utah} \\
Salt Lake City, UT, USA \\
atharv.belsare@utah.edu
% email address or ORCID
}
\and
\IEEEauthorblockN{Zohre Karimi$^*$}
\IEEEauthorblockA{\textit{University of Utah} \\
Salt Lake City, UT, USA\\
zohre.karimi@utah.edu
% email address or ORCID
}
\and
\IEEEauthorblockN{Connor Mattson$^*$}
\IEEEauthorblockA{\textit{University of Utah} \\
Salt Lake City, UT, USA \\
c.mattson@utah.edu
% email address or ORCID
}
\and
\IEEEauthorblockN{Daniel S. Brown}
\IEEEauthorblockA{\textit{University of Utah} \\
Salt Lake City, UT, USA \\
daniel.s.brown@utah.edu
% email address or ORCID
% }
% \and
% \and
}}

\maketitle

%% Equal Contribution Footnote
\def\thefootnote{$*$}\footnotetext{Equal contribution.}\def\thefootnote{\arabic{footnote}}
\def\thefootnote{}

\begin{abstract}
A fundamental challenge of shared autonomy is to use high-DoF robots to assist, rather than hinder, humans by first inferring user intent and then empowering the user to achieve their intent.
Although successful, prior methods either rely heavily on \textit{a priori} knowledge of all possible human intents or require many demonstrations and interactions with the human to learn these intents before being able to assist the user.
We propose and study a zero-shot, vision-only shared autonomy (VOSA) framework designed to allow robots to use end-effector vision to estimate zero-shot human intents in conjunction with blended control to help humans accomplish manipulation tasks with unknown and dynamically changing object locations.
To demonstrate the effectiveness of our VOSA framework, we instantiate a simple version of VOSA on a Kinova Gen3 manipulator and evaluate our system by conducting a user study on three tabletop manipulation tasks. 
The performance of VOSA matches that of an oracle baseline model that receives privileged knowledge of possible human intents while also requiring significantly less effort than unassisted teleoperation. 
In more realistic settings, where the set of possible human intents is fully or partially unknown, we demonstrate that VOSA requires less human effort and time than baseline approaches while being preferred by a majority of the participants. Our results demonstrate the efficacy and efficiency of using off-the-shelf vision algorithms to enable flexible and beneficial shared control of a robot manipulator. Code and videos available here: \href{https://sites.google.com/view/zeroshot-sharedautonomy/home}{https://sites.google.com/view/zeroshot-sharedautonomy/home}
\end{abstract}


\section{Introduction}
% This document is a model and instructions for \LaTeX.
% Please observe the conference page limits. 

% \footnotetext{Equal Contribution}

% Task
Assistive robot arms have the potential to enable millions of people worldwide with disabilities to perform everyday living tasks~\cite{taylor2018americans,argall2018autonomy,mitzner2018closing,losey2022learning}. Imagine a user with a disability who has access to a wheelchair-mounted robot arm. While they can teleoperate this arm via a joystick, 
%Robot teleoperation enables humans to directly control the motion of their robots. Though direct teleoperation matches the human's inputs exactly, 
the dimensionality required to control high-DoF robots makes teleoperation challenging~\cite{pierce2012data, meeker2018intuitive, jeon2020shared}.
%, who may be accustomed to providing input to systems with 2 or 3 dimensions at most. 
% A long-standing desire of teleoperation is to retain robot precision while lowering the expertise and burden required to pilot these systems. 
% Figure
\begin{figure}[t]
     \centering    \includegraphics[width=0.80\linewidth]{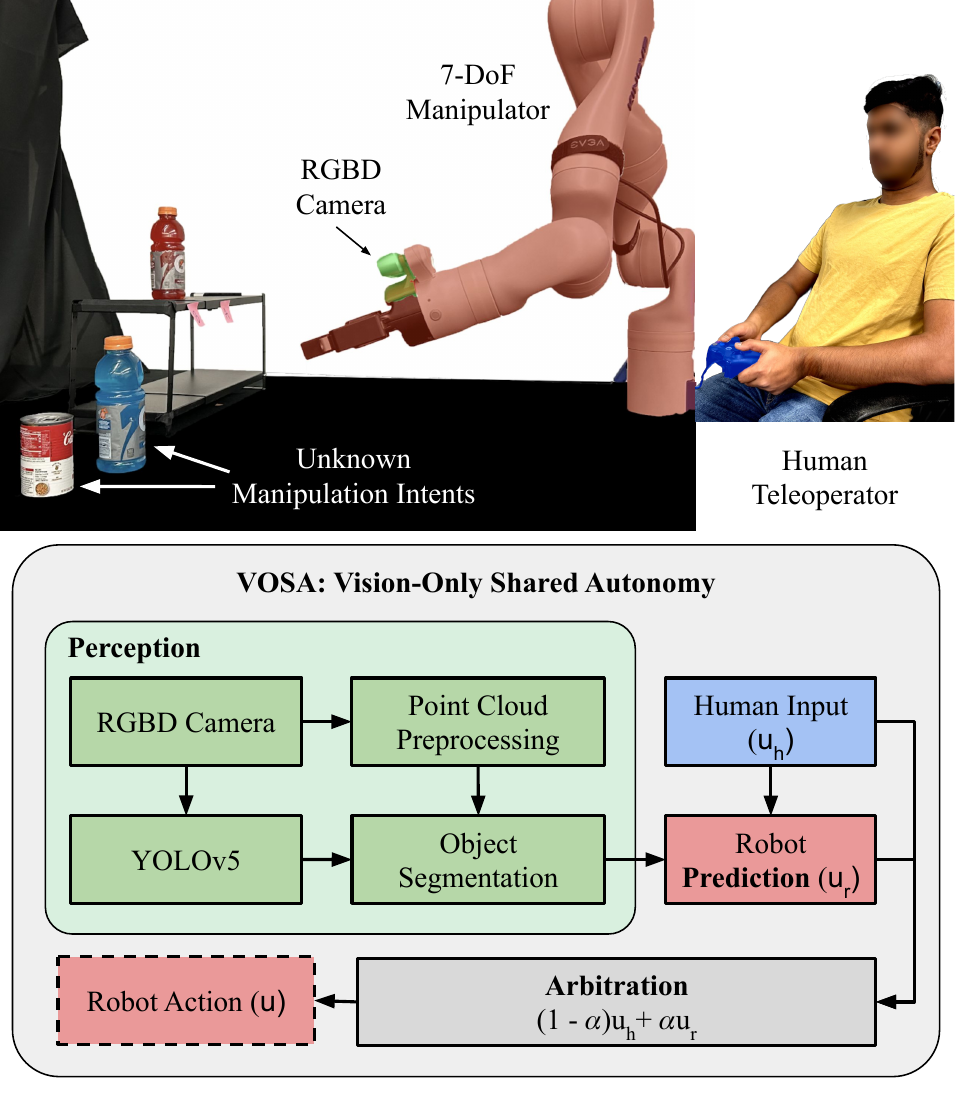}
    \caption{\textbf{Vision-Only Shared Autonomy (VOSA)} combines the benefits of shared autonomy and robot perception to generalize out-of-the-box to new scenes by dynamically \textbf{perceiving} all possible intents, \textbf{predicting} the human's desired intent, and \textbf{arbitrating} the human and robot control actions.} 
    % \caption{\textbf{Vision-Only Shared Autonomy (VOSA)} combines the benefits of shared autonomy and computer vision to generalize out-of-the-box to new scenes by dynamically representing the set of possible human intents using segmented point clouds from a single wrist-mounted RGBD camera. }
    \label{fig:teaser}
\end{figure}
To provide teleoperation assistance, many researchers have looked to shared autonomy~\cite{selvaggio2021autonomy, kim2024taxonomy, udupa2023shared}. Shared autonomy provides a framework for blending human intent with robot inference by prioritizing human control when the robot has low confidence in user intent and prioritizing artificial control when the robot has high confidence. Particularly exciting progress has been shown when shared autonomy systems are represented as prediction-then-arbitration~\cite{dragan_formalizing_2012}. However, one of the main limitations of prior work is that it typically assumes access to a predefined set of possible user intents~\cite{admoni2016predicting,gopinath_human---loop_2017,brooks2019balanced,dragan_formalizing_2012,jain2019probabilistic,javdani2018shared,jonnavittula2021know,newman2022harmonic,nikolaidis2017human}.

Consider someone using an assistive robot arm to help their roommate or partner put away groceries (see Figure~\ref{fig:teaser}). In this and many other real-life settings, the objects that the user wishes to manipulate may be initially unknown and may be non-stationary. For example, the roommate may be unloading grocery items onto the table, and the user of the assistive arm may be putting them on a shelf. The robot will typically not know the type or location of target objects and the object locations may be constantly changing---the roommate may put some of the objects away themselves or move objects to make room for others. Despite this, most prior work in shared control assumes access to a known set of user intents (often in the form of predefined object types and locations). By contrast, we propose that shared autonomy should leverage computer vision to enable assistive robot manipulation via shared control that can work in an out-of-the-box fashion in these kinds of dynamic situations. 
While there has been some work that generalizes shared autonomy to unknown intents and tasks, prior work requires access to high-quality demonstrations to learn from, which can be difficult to obtain and limits the scalability and applicability of these approaches~\cite{zurek2021situational,jonnavittula_sari_2023,karamcheti_learning_2021}. 
% By contrast, we study the effectiveness of a simple zero-shot approach to shared autonomy that requires no pretraining nor learning from demonstrations.
In contrast, we explore a simple zero-shot shared autonomy approach that needs no pretraining or demonstrations.

The main contributions of our work are the introduction and instantiation of a new framework for zero-shot shared autonomy and an evaluation of this approach via a user study. We introduce Vision-only Shared Autonomy (\textbf{VOSA}), a novel, simple, and effective method of shared control for tabletop manipulation tasks that do not require a set of known possible intents \textit{a priori} and achieves high performance out-of-the-box with no need for online or offline learning from demonstrations.
% Idea
VOSA builds upon the success of the prediction-then-arbitration framework~\cite{dragan_formalizing_2012} by formally incorporating intent perception into a \textit{perception-then-prediction-then-arbitration} framework. This framework aids humans in teleoperating robots by \textit{predicting} the set of discrete manipulation intents in a scene instantaneously using a camera affixed to the end-effector of a robot arm and then \textit{arbitrating} control between the human's input and the robot's prediction. To the best of our knowledge, we are the first to consider zero-shot intent inference for shared autonomy using only end-effector vision.

% Proof
We evaluate both the objective and subjective performance of VOSA in a user study (N=18 users) with three different manipulation tasks. We compare VOSA's performance to a shared autonomy baseline~\cite{gopinath_human---loop_2017}, which has oracle knowledge of possible intents, and to a direct teleoperation baseline, the current industry standard~\cite{jaco}.
%and compare the performance of VOSA to a shared autonomy baseline~\cite{gopinath_human---loop_2017}, which receives oracle knowledge of possible intents, and to a direct teleoperation baseline which is the current industry standard~\cite{jaco}. We
We show that even without oracle knowledge of human intent, VOSA is competitive with the oracle model and significantly outperforms direct teleoperation in terms of human input and task duration.
% We show that even without oracle knowledge of human intent, VOSA is competitive with the baseline oracle model and significantly outperforms direct teleoperation as measured by human input and task duration.

We also show that shared autonomy baseline methods fail when the set of possible human intents is misrepresented or unknown, causing large amounts of user frustration and hindered performance. By contrast, VOSA achieves high zero-shot performance on all tasks, including tasks with non-stationary objects due to the ability to dynamically update possible intentions at runtime.
We find that users prefer VOSA over baselines in realistic settings where the baselines must deal with ambiguous or only partially known user intent and rate VOSA as a significant improvement over baselines in terms of task recognition and robot helpfulness.

\section{Related Work}

A spectrum of shared autonomy frameworks has been developed to enhance collaboration between humans and robots across various applications, ensuring more intuitive and efficient human-robot interactions.

Policy blending, the arbitration approach used in this paper, is a popular assistive formalization that combines user's intent with robot's prediction of those intentions ~\cite{dragan_formalizing_2012,jain2019probabilistic,zurek2021situational,newman2022harmonic,jonnavittula_sari_2023}. Accurately predicting a user intention is one of the core challenges of shared autonomy~\cite{selvaggio2021autonomy,udupa2023shared}.
Prior work mainly focuses on shared autonomy with predefined goals or intents~\cite{gopinath_human---loop_2017,brooks2019balanced,dragan_formalizing_2012,jain2019probabilistic,javdani2018shared,jonnavittula2021know,newman2022harmonic,nikolaidis2017human}. Other work seeks to learn models of human intent from demonstrations~\cite{reddy_shared_2018,jonnavittula_sari_2023,zurek2021situational,mehta2022learning,yoneda_noise_2023} or use natural language to disambiguate user intent~\cite{nikolaidis2018planning,mo2023towards,yow_shared_2024}. In our work, we focus on zero-shot shared autonomy where the intents are not predefined, and we do not require the user to teach the robot their intent via demonstrations or language.

We study how we can enable the robot to dynamically infer manipulation intents in its environment using a camera mounted on the end-effector. There is an extensive set of prior works on enabling vision-based grasping for assistive robots (see ~\cite{bai2020object,du2021vision} for comprehensive surveys), 
many of which utilize stereo or depth cameras that observe the task scene from a fixed perspective (eye-to-hand)~\cite{loconsole2014robust, zhang2017intention, brooks2019balanced, arrichiello2017assistive, zeng2017closed, pitzer2011towards}. Some work has utilized end-effector vision (eye-in-hand) for perception in shared autonomy, but relies on additional external cameras to fully capture scene information~\cite{gualtieri2017open, elarbi2013eye}. Other work has studied the use of end-effector vision as the sole perception input for assistive grasping and motion planning, but assumes that a desired object was selected by the human prior to autonomous assistance and that the scene remains unchanged at runtime~\cite{kim2011motion, kim2009empirical,remazeilles2008sam}, effectively simplifying shared autonomy to robot control via visual servoing.
Surprisingly, we are not aware of any studies that have evaluated the efficacy of using the perception capabilities of a wrist-mounted camera to enable zero-shot intent inference in combination with realtime command arbitration as we do in this paper.

Other work on assistive control replaces joystick controls with screen-based selection processes~\cite{quintero_vibi_2015, elarbi2013eye, remazeilles2008sam, kim2011motion, pitzer2011towards, hagenow2024system}, natural language commands~\cite{rajapakshe2024synergizing, cui2023no}, eye gaze~\cite{zeng2017closed, aronson2020eye}, wearable devices~\cite{padmanabha2024independence}, or brain-computer interfaces~\cite{xu_shared_2019, zhang2017intention, arrichiello2017assistive, zeng2017closed} to seek to enable users to teleoperate robots. We focus on joystick teleoperation, but our framework naturally extend to other interfaces. 
% Other work uses force sensing and point cloud data to assist teleoperation for ultrasound imaging~\cite{yang_human-robot_2023} or integrates visual state representations with shared autonomy by leveraging pretrained object detectors and conditional autoencoders trained from demonstrations~\cite{karamcheti_learning_2021}.
In contrast to prior work, we propose a new framework for shared autonomy with the goal of vision-based intent recognition to enable robots to adapt to new and unseen environments without requiring pre-programmed manipulation intents or demonstrations by leveraging scene understanding solely from an end-effector camera. %This approach eliminates the need for static depth cameras, easing the path to possible future deployment and enabling real-time updates of possible human intents. 
This approach removes the need for static depth cameras, simplifying future deployment and allowing real-time updates of human intents.

% Fontaine et al. focus on utilizing the Quality Diversity (QD) method, particularly MAP-Elites, for creating diverse interaction scenarios between humans and robots. This approach aims to explore a wide range of failure scenarios by examining different environmental conditions and human actions to evaluate shared autonomy algorithms better\cite{fontaine_quality_2021}. \\

% \subsection{Maintaining the Integrity of the Specifications}

% The IEEEtran class file is used to format your paper and style the text. All margins, 
% column widths, line spaces, and text fonts are prescribed; please do not 
% alter them. You may note peculiarities. For example, the head margin
% measures proportionately more than is customary. This measurement 
% and others are deliberate, using specifications that anticipate your paper 
% as one part of the entire proceedings, and not as an independent document. 
% Please do not revise any of the current designations.

\section{Problem Statement}

% \textbf{TODO: Explicitly formalize what it means to be ``zero-shot"}
\begin{figure*}[h]
    \centering
    \begin{subfigure}[t]{0.12\textwidth}
        \centering
        \includegraphics[width=\linewidth]{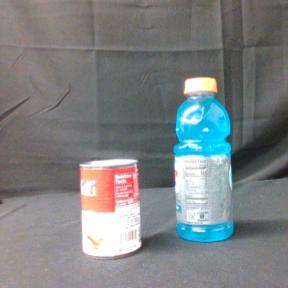}
        \subcaption{}
    \end{subfigure}%
    ~ 
    \begin{subfigure}[t]{0.12\textwidth}
        \centering
        \includegraphics[width=\linewidth]{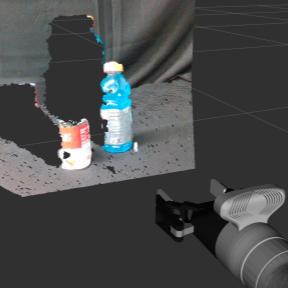}
        \subcaption{}
    \end{subfigure}
    ~
    \begin{subfigure}[t]{0.12\textwidth}
        \centering
        \includegraphics[width=\linewidth]{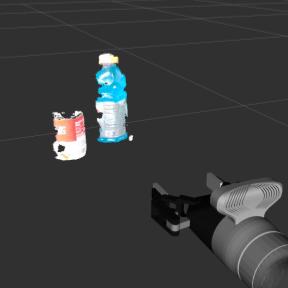}
        \subcaption{}
    \end{subfigure}
    ~
    \begin{subfigure}[t]{0.12\textwidth}
        \centering
        \includegraphics[width=\linewidth]{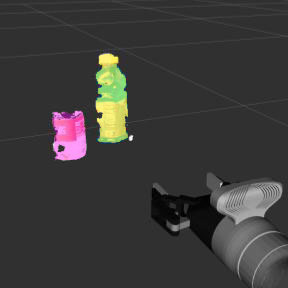}
        \subcaption{}
    \end{subfigure}
    ~
    \begin{subfigure}[t]{0.12\textwidth}
        \centering
        \includegraphics[width=\linewidth]{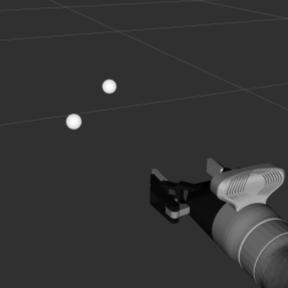}
        \subcaption{}
    \end{subfigure}
    \caption{\textbf{A Simple Instantiation of VOSA Perception.} (a) The 2D raw RGB scene
    %, featuring a soup can (left) and a sports drink (right), 
    from the perspective of a camera mounted at the robot's end-effector. 
    (b) The 3D RGB + Depth (RGBD) scene as a point cloud in Simulation. 
    (c) Point cloud preprocessing filters out the table and background. (d) $k$-means clustering is used to classify points to an object. The value of $k$ is obtained from a pretrained YOLOv5~\cite{redmon2016look} model. (e) Centroids of the point cloud clusters represent possible human intents. }
\end{figure*}
In the previous section, we explored various approaches to shared autonomy. Many of these studies, however, assume conditions that may not be feasible in many real-world applications, such as the availability of additional external cameras, fiducial markers on objects of importance, or reliance on predefined sets of possible user intents. In contrast, our focus is on implementing and evaluating a shared autonomy framework that works on out-of-the-box manipulators with a single wrist mounted depth camera.
%that enables shared autonomy to b deployed in more general scenarios, where the robot has no prior knowledge of the objects' geometry or locations.
Additionally, we aim to explore a zero-shot method, where the robot can be deployed without any \textit{a priori} knowledge, pretraining or demonstrations, that has the potential to generalize across a wide range of manipulation scenarios. As a step towards these goals, we formalize a general framework to address the challenge of zero-shot shared autonomy.
%In this paper, we formalize a general framework to tackle the challenge of zero-shot shared autonomy.}
% Thus we seek to formalize \st{provide} a general solution to address the problem of zero-shot shared autonomy.

% \edit{Inspired by previous formulations of human-robot interaction~\cite{parasuraman2000model, pitzer2011towards}, we consider a shared autonomy system that consists of three high-level major components (depicted in Figure~\ref{fig:teaser}): Perception, Arbitration, and Prediction. Perception is used to identify manipulation intents in an environment. The system does not know the user's true intent \textit{a priori}. Rather, the robot uses its own observations and human inputs to reason about potential intents that exist in a continuous region of the workspace. This approach allows for a dynamic interaction where the robot can adapt to new information and user behaviors effectively. Arbitration is used to make a decision that balances control between the human input and the robot's autonomy. Finally, Prediction is used to anticipate both user intentions and environmental changes, enabling adjustments to the changes in the intent.}

Inspired by previous formulations of human-robot interaction~\cite{parasuraman2000model, pitzer2011towards}, we consider a shared autonomy system that consists of three high-level  components (depicted in Figure~\ref{fig:teaser}): Perception, Prediction and Arbitration. Perception is used to identify manipulation intents in an environment. Prediction uses the set of possible intents identified by Perception, along with the robot's state and current human inputs, to reason over these intents and predict the user's true goal. This step is critical in shared autonomy, as the system does not know the user’s true intent \textit{a priori}, and the intent may change during execution. Arbitration balances control between the human inputs and the robot’s predictions to make a decision about which action to perform.%informed decision.

Thus, what sets our problem setting apart from prior work in shared autonomy is our focus on shared autonomy using a single, end-effector mounted, camera and no access to a predefined set of possible user intents. The primary motivation of our work is to explore how to handle scenarios where manipulation intents are not predefined and may change dynamically as objects enter or leave the robot’s workspace. We implement a simple yet effective approach to address this challenge and evaluate its effectiveness across various manipulation tasks.
%The main motivation of our work is to explore how to address scenerios where manipulation intents are not predefined and may change over time due to objects entering and leaving the robot's workspace. We implement a simple, but effective, approach to study this problem and evaluate its efficacy on different manipulation tasks.

% \edit{The main motivation of our work is to study how to close the gap between nvironments where the possible manipulation intents are unknown beforehand and may change over time due to objects entering and leaving the robot's workspace. We implement a simple, but effective approach to study this problem and evaluate its efficacy on different manipulation tasks.}

Following previously defined notation for shared autonomy~\cite{javdani_shared_2015, dragan_formalizing_2012}, let $X$ represent the robot's continuous state space (e.g., position and velocity), and let $U_r$ represent the continuous space of robot actions (e.g., joint velocities, torques). The dynamics of the robot can therefore be modeled as the transition function $T: X \times U_r \to X$. \textit{Teleoperation} uses human inputs, $u_h \in U_h$ to directly command actions for the robot. Finally, we assume that there is a finite set of visually perceptible user intents, $G \subset \mathcal{G}$, which may consist of objects to interact with, or surfaces to place objects on, to name a few examples. In this paper, we explore \textit{Shared Autonomy}, where the robot infers the human's desired intent and proposes an action, $u_r \in U_r$ for the robot to take. Then, an arbitration function, $A: U_h \times U_r \times \mathcal{G} \times X \to U_r$, outputs a blended robot action as a function of the human's input, the robot's predicted action, and the state of the scene. In contrast to prior work, we extend shared autonomy to scenarios where $G$ is not known \textit{a priori} and may even change, requiring additional inference.

\section{Method}

\begin{figure*}[t]
    \centering
    \begin{subfigure}[t]{0.25\textwidth}
        \centering
        \includegraphics[width=\linewidth]{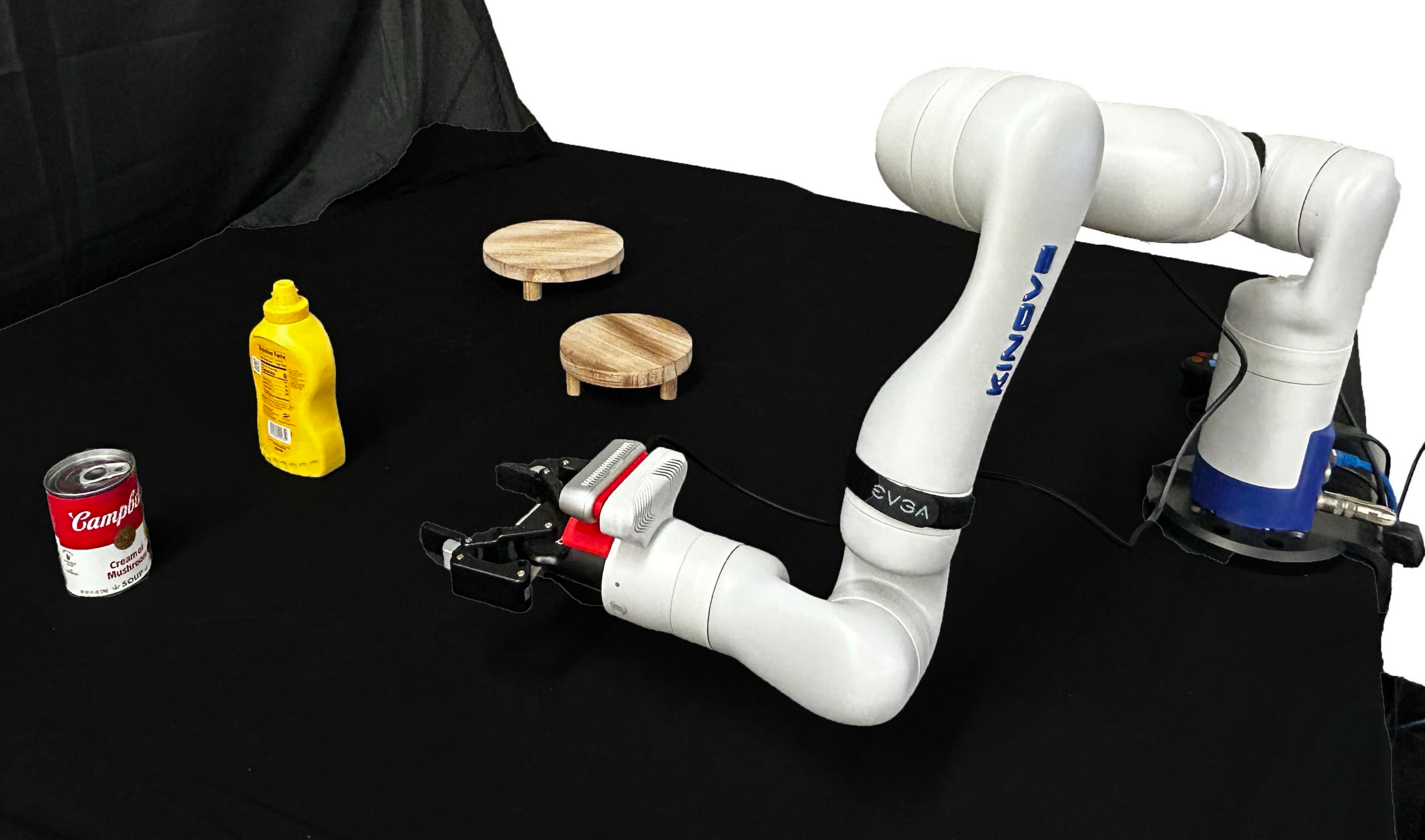}
        \subcaption{Pick and Place}
        \label{Pick and place}
    \end{subfigure}%
    ~ 
    \begin{subfigure}[t]{0.25\textwidth}
        \centering
        \includegraphics[width=\linewidth]{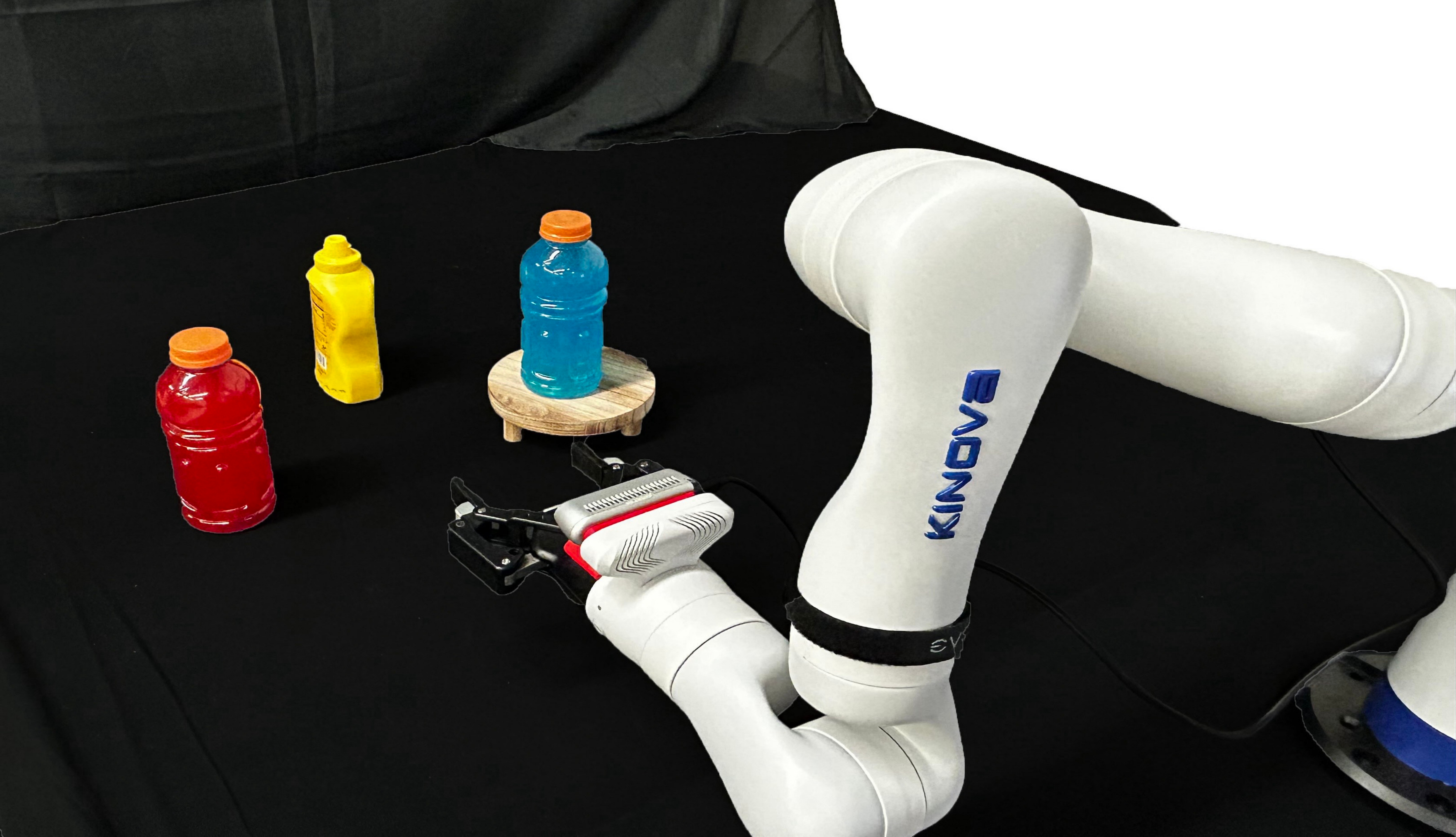}
        \subcaption{Deceptive Grasping}
        \label{Deceptive reaching}
    \end{subfigure}
    ~
    \begin{subfigure}[t]{0.25\textwidth}
        \centering
        \includegraphics[width=\linewidth]{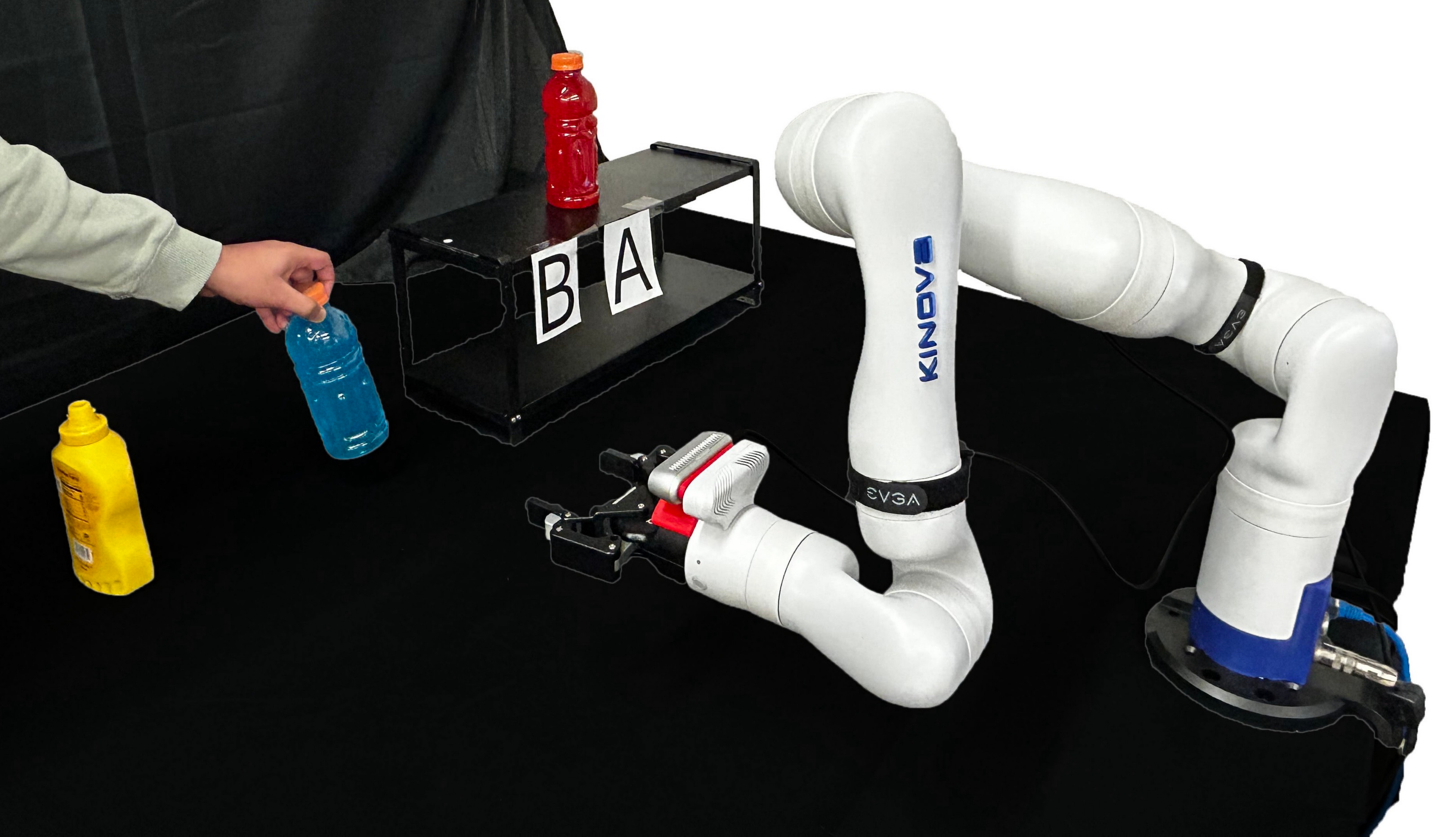}
        \subcaption{Shelving}
        \label{shelving}
    \end{subfigure}
    \caption{\textbf{User Study Tasks.} (a) Pick and Place task where users were asked to relocate two objects in the scene to indicated placement intents (wood pedestals). (b) Deceptive Grasping task where the robot must reach in between two other objects to grasp the target object. (c) Shelving task where the robot helps a human teammate stock a shelf with sports drinks.}
    \label{fig:experiment-setup}
\end{figure*}

In this section, we describe our proposed framework, \textit{Vision-Only Shared Autonomy (VOSA)}, for a shared autonomy system that uses Computer Vision (CV) to dynamically approximate the set of potential user intents in the environment at runtime.
Our framework is designed to function out-of-the-box as a means of zero-shot assistive manipulation. While our proposed implementation represents a simple instantiation of VOSA, it demonstrates the effectiveness of combining shared autonomy with off-the-shelf CV algorithms to achieve real-time inference in dynamic environments. 
% \subsection{Point Cloud Preprocessing}
\subsection{Perception}
% The first step for the robot to perceive and interact with the environment is using sensors, and among the various sensors, cameras play an essential role by providing both color and depth information
In VOSA, the information extracted from the environment comes from the RGBD data obtained from a camera affixed to our robot's end-effector. 
% The camera returns a 2D array where each pixel contains a color value (RGB) and a depth measurement , approximating the distance in the world space from the corresponding point to the camera's lens. \edit{We retrieve this information from the camera as a point cloud and transform it to be at the robot's world frame.} 
The camera captures a 2D array of pixel data, each with RGB color values and depth measurements, which is processed into a point cloud.
We preprocess the point cloud to remove points that should not be considered in the potential intent set, such as tables and walls.
%Also, we remove points from the point cloud that should not be considered in the potential intent set, such as tables and walls. 

% We retrieve this information as a point cloud where the RGBD values are transformed to a set of size $N$ where elements of the set are points $p_{c} = (x_{c}, y_{c}, z_{c}, r, g, b)$, where the $c$ subscript indicates that points are in the reference frame of the camera. Therefore, the point cloud, $P$, with respect to the camera frame is $P_{c} = \{p_{i, c} \ | \ i \in 1, 2, ..., N\}$.
% Because our camera is mounted on the end-effector, the reference frame of the camera is dependent on the state of the robot. Therefore we transform our point cloud to a world reference frame, $w$, defined at the base of the robot, using the transformation matrix $^{w}T_{c}$, so we have $P_{w} =$ $^{w}T_{c} P_{c}$. 

% Our work assumes that objects of interest lie in a constant axis-aligned rectangular prism, $\mathcal{G}$, defined by upper and lower limits on $x_w$, $y_w$, and $z_w$. By only considering points within this region, $P_{w}' = P_w \cap \mathcal{G}$, 

% \subsection{Intent Inference via Point Cloud Segmentation}

After preprocessing the point cloud, we need to consider how to identify objects in the scene and segment them into individual point clouds. Because we assume no privileged information about the set of user intents, our method must not only infer where these objects are but also how many objects are represented in the scene. A simple way to achieve object segmentation is to use unsupervised clustering algorithms to associate dense clusters of points with the same class label. Our method utilizes the $k$-Means algorithm, with the value of $k$ determined by the number of objects detected in the scene using a pretrained YOLOv5~\cite{redmon2016look} model. YOLOv5 is well-suited for this framework as we only use the number of objects it detects to correctly identify the number of objects, not the classification labels. In this work, we assume even if the labels are inaccurate, YOLOv5 still provides the correct number of objects. The centroid of each cluster forms the set of possible manipulation intents, $G \subset \mathcal{R}^3$.

One immediately apparent disadvantage of using only end-effector vision for shared autonomy is camera hardware constraints. In particular, many depth cameras that rely on stereo vision deteriorate in quality as objects get very close to the camera. To address this in our framework, we assume that $G$ is only updated when the camera is viewing the scene from within the manufacturer-specified viewing distance.

\subsection{Prediction}

After using CV to determine the intent set, $G$, the robot must predict the next autonomous action that it would take based on the information in the scene. For each potential intent, $g \in G$, and end-effector position at time $t$ as $x(t)$, a robot action, $u_{r, g}(t)$, is generated to simulate a unit step in Cartesian space toward the potential intent $g$. 
\begin{equation}
u_{r,g}(t) = \frac{g - x(t) }{\lVert g - x(t) \rVert} 
\label{eq:ur}
\end{equation}
Following prior work~\cite{gopinath_human---loop_2017}, each potential intent is assigned a confidence level based on the Euclidean distance $d(g)$ between the end-effector position and $g$ at time $t$, 
%with $c(t)$ ranging between 0 and 1. $c(t)$ is computed as: 
\begin{equation}
c(t, g) = w_{1} \cdot (u_{h}(t) \cdot u_{r, g}(t)) + w_{2} \cdot e^{-d(g)}
\end{equation}
This equation combines two components: The first reflects the agreement between the command provided by the user and the command generated by the robot, while the second one measures the proximity of the end-effector to the potential intent. The importance of these two factors is adjusted through weights, $w_{1}$ and $w_{2}$, which we tuned based on a small pilot study and found that $w_1=0.3, w_2 =0.7$ worked best across all of our experimental tasks, %considered three different pairs of weights $(w_{1},w_{2})$ which were $(0.4,0.6)$,$(0.2,0.8)$, and $(0.3,0.7)$. We tried to see the robot performance in completing some reaching objects task with different, challenging object configuration in the scene. Between them, $(0.3,0.7)$ works better in term of 
in terms of ensuring good robot assistance while enabling the human to correct the robot if needed.
After calculating the confidence for each potential intent, the robot selects the action and corresponding confidence level of the most confident intent, denoted as $u_r(t)$ and $c(t) = \arg \max_{g\in G} c(t, g)$ respectively.

\subsection{Arbitration}

\begin{figure}
    \centering
    \includegraphics[width=0.6\linewidth]{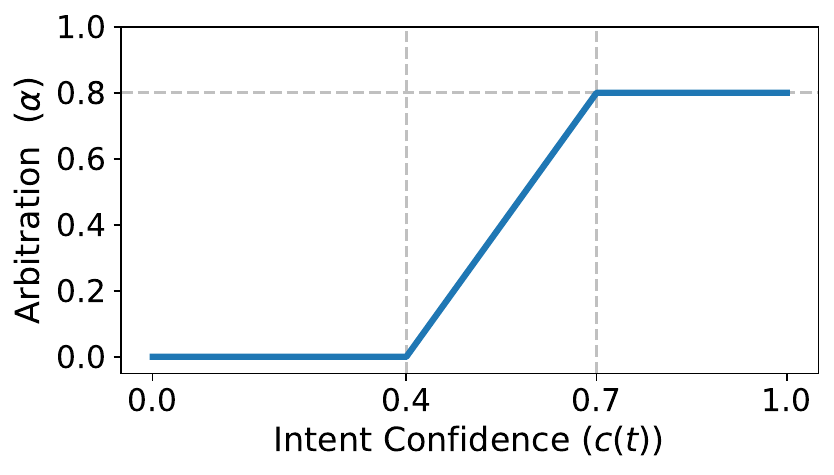}
    \caption{The arbitration function increases the influence of the robot's command ($\alpha$) as the robot's confidence $c(t)$ in inferring the human's true intent increases.}
    \label{fig:Arbitration}
\end{figure}

In modeling the state of the system, $x(t)$, at $t$, we incorporate dual control inputs: $u_h(t)$  and $u_r(t)$ which represent the command generated by the human user and robot's policy respectively, as shown in Figure~\ref{fig:teaser}. To arbitrate between the human and robot control commands, we follow the popular linear blending approach~\cite{dragan_formalizing_2012} to generate the control command $u(t)$, that the robot will execute: 
\begin{equation}
u(t) \gets (1 - \alpha) \cdot u_h(t) + \alpha \cdot u_r(t)
\label{eq:arbitration}
\end{equation}
% In this design, the function $\boldsymbol{\beta}(\cdot)$, which meditates between the interaction between robot and human control signals, operates as a linear blending function. 
% \[
% \boldsymbol{\beta}(u_h(t),u_r(t))= (1 - \alpha) \cdot u_h(t) + \alpha \cdot u_r(t))
% \]
Here $\alpha$ arbitrates control between the human and the robot. If $\alpha=0$, the human is in full control; and if $\alpha = 1$, the robot is in full control. Following prior work~\cite{dragan_formalizing_2012,gopinath_human---loop_2017}, we dynamically set $\alpha$ as a function of the robot's confidence, $c(t)$, in its prediction of the user's intent as shown in Figure~\ref{fig:Arbitration}. Building on previous works~\cite{kim2011autonomy, you2012assisted, javdani_shared_2015, Gombolay-RSS-14, 5941028}, which highlighted the user's preference for maintaining control and correcting the robot's action, We limit the maximum arbitration to $\alpha=0.8$, ensuring that the human always has some control over the actions taken by the robot, but when needed they can pass control to the robot by not providing any input.
%but they still have the option to do nothing and put their trust in the robot to complete the task for them.} 
%Along the x-axis is the robot confidence at time $t$, denoted by $c(t)$. 
% This confidence level, $c(t)$, is evaluated at every moment when the human issues a non-zero control input, $u_h(t)$.

To determine $\alpha$, we use the highest confidence, $c(t)$, as depicted in Figure \ref{fig:Arbitration}. Subsequently, $\alpha$ is employed in Equation \ref{eq:arbitration} to blend human and robot commands. This process of calculating intent confidence and arbitration values occurs continuously during runtime, allowing for real-time adjustments based on the robot's movement and any modifications to the set of inferred possible intents $G$ based on its real-time perception.

%making it easy for the user to control the robot and preventing robot from going to wrong goal or sucking in some object configuration that can not decide which object is the user intent. These weights can be tuned depending on the significance you place on certain terms to enhance the effectiveness of the final command issued to your robot. Eventually, we used this pair for our experiments.}

\subsection{State Transitions} \label{sec:transitions}

In our experiments, we focus on tabletop manipulation of objects where the robot needs to pick and place multiple objects. To simplify our user study design, we assume access to a known set of \textit{object placement intents} (e.g., the robot knows the unoccupied locations on the shelves). In a real deployment, the robot would likely already know where objects could be placed (a caregiver with AR tags could specify them) but will not know what objects should be placed there nor where these objects may originate from. These locations could be easily learned or identified via computer vision in future work. 

Since our tasks involve two sets of intents (objects that need to be grasped and placement locations), we model the system as a state machine with four distinct states:
    \textbf{(1) Object Sensing:} The robot processes RGBD data from the camera to detect possible manipulation intents, $G$, in the form of graspable objects. \textbf{(2) Grasping:} After identifying $G$, arbitration continues until the robot's end-effector is positioned at the intended object. The state transitions once the user closes the gripper, signaling satisfaction with the grasp. \textbf{(3) Placing:} The robot then infers where the user intends to place the object. The user can choose to open the gripper at any time, signaling the end of a placement. \textbf{(4) Active Sensing:} The robot reorients itself to a home position, allowing the camera to gain space for new object-sensing, restarting the process.

    % In this initial state, the robot continually processes the RGBD data from the camera to identify and locate objects of interest that may correspond to user intents ($G$). This state is active until the camera moves outside of its pre-specified sensing range ($\ge45$cm away from the active sensing home position).
    
    % Using the frozen approximation for $G$, arbitration continues as normal until the end-effector is positioned at the user's intended object. The transition to the next state occurs once the human closes the gripper and provides a non-zero input for the robot, indicating that the user is satisfied with the grasp.

    % With the object grasped, the Placing state continues shared control but now aims to infer intent over the object placement locations, rather than object locations. The human releases the object by opening the gripper and indicating satisfaction with the placement by manually triggering the transition to the subsequent state.

    % Finally, Active Sensing involves the system returning the robot's end-effector to the home position so that the camera is reoriented with enough space to begin a new object-sensing phase.

These states allow the system to alternate between identifying objects to grasp and determining placement locations, ensuring smooth transitions between object sensing, grasping, placing, and reorienting.
\section{Experiments}
\begin{figure*}
    \centering
    \begin{subfigure}[t]{0.40\textwidth}
        \centering        \includegraphics[width=\linewidth]{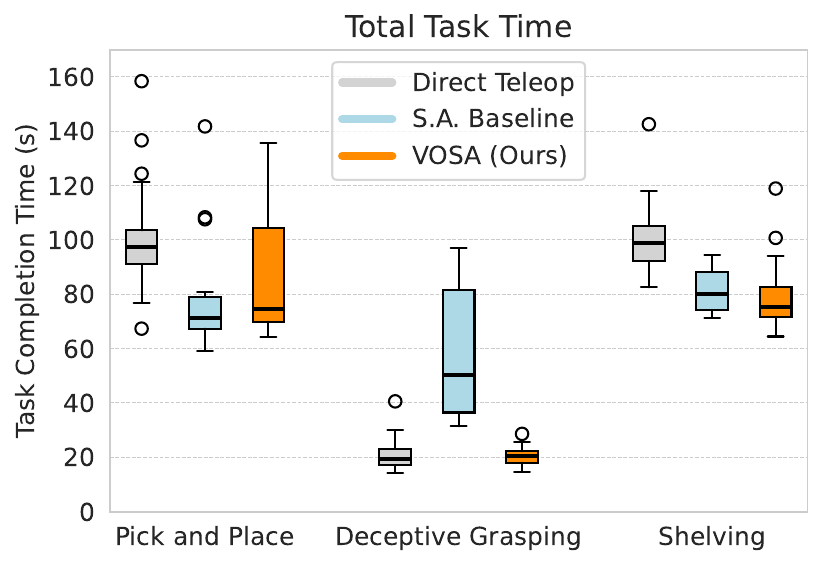}
        \subcaption{}
        \label{fig:task-time}
    \end{subfigure}%
    % \hfill 
    \hspace{5mm}
    \begin{subfigure}[t]{0.40\textwidth}
        \centering    \includegraphics[width=\linewidth]{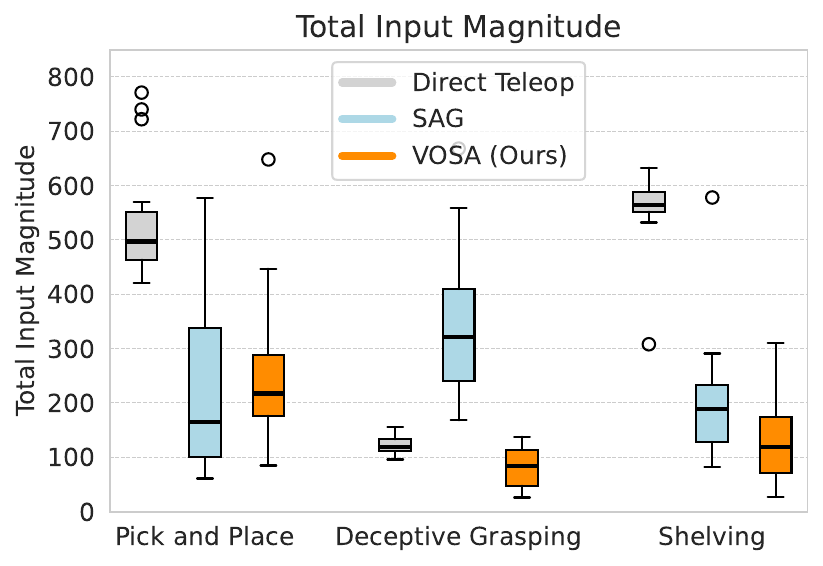}
        \subcaption{}
        \label{fig:total-effort}
    \end{subfigure}
    ~
    % \begin{subfigure}[t]{0.32\textwidth}
    %     \centering
    %     \includegraphics[width=\linewidth]{media/prefs.pdf}
    % \end{subfigure}
    \caption{\textbf{Quantitative Results.} (a) Task Completion Time and (b) Input Magnitude for a user study of 18 total users. Subjects were exposed to three shared autonomy paradigms: direct teleoperation, a shared autonomy baseline with known intents (SAG), and Vision-only shared autonomy (VOSA). Note that if SAG is instantiated without oracle goal information, it functionally reduces to direct teleoperation, whereas VOSA is able to adapt on-the-fly and infer new intents at runtime. In cases where direct teleoperation is burdensome (pick and place, shelving) and cases where intents are not correctly specified prior to the task (shelving, deceptive grasping), VOSA provides a zero-shot assistance paradigm that reduces both burden and intent uncertainty.}
    \label{fig:quant-results}
\end{figure*}
We conducted a user study (N=18 users) using the Kinova Gen3 7-DoF robotic arm with a Realsense D435i camera attached to the end-effector. The users were provided an Xbox 360 joystick controller to control the arm. The users controlled the robot's end-effector using just the left joystick and operated in two different control modes: the x-y-axis control mode and the z-axis control mode. The control interface for the robot was designed to resemble the joystick controls commonly used to control assistive robot arms~\cite{jaco,gopinath_human---loop_2017,DBLP:journals/corr/abs-1909-09674}. 
% Code to reproduce all of our experiments is available here: \textit{\url{https://github.com/aria-lab-code/VOSA}}
%\textit{[link removed for anonymous submission]}.

\subsection{Baselines and Treatments}

Each task in the user study evaluated three different treatments: (A) Direct Teleoperation (\textbf{Teleop}) by the user using the provided Xbox 360 joystick; (B) Shared Autonomy with Known Goals (\textbf{SAG})~\cite{gopinath_human---loop_2017}, where the shared autonomy is computed using access to privileged knowledge of possible user intents---we study cases where this privileged information is perfect and more realistic settings where it is incomplete or inaccurate; and (C) \textbf{VOSA}, which is the same as SAG, except our method does not receive any privileged information related to the object locations and must rely on vision and active sensing to understand possible user intents. SAG and VOSA were provided with complete and accurate knowledge of object placement locations as mentioned in \ref{sec:transitions}.
The order of treatments given to the users in each task was randomized, and the users were not made aware of which treatment they were given. They were told that in some treatments the robot would help them with the task.
For all the treatments, we ensured consistency across parameters to maintain uniformity in the experimental conditions. These parameters included the command arbitration function, the method used for estimating intent, and the speed of the end-effector (5 cm/s). 

\subsection{Hypotheses}
We test the following hypotheses: \textbf{H1:} VOSA assists users in accomplishing tasks in less time than Teleop and is not inferior to SAG when SAG has accurate oracle knowledge of all possible human intent. \textbf{H2:} VOSA requires less human effort than Teleop and is not inferior to SAG when SAG has accurate oracle knowledge of all possible human intent. \textbf{H3:} VOSA is superior to SAG when SAG has incomplete/inaccurate knowledge of possible human intent. \textbf{H4:} Participants will qualitatively indicate that VOSA is better at recognizing the objective of tasks compared to SAG during tasks where SAG has incomplete/inaccurate knowledge of possible human intent. \textbf{H5:} Participants will qualitatively indicate that VOSA is more helpful than SAG during tasks where SAG has incomplete/inaccurate knowledge of possible human intent.

\subsection{Task Description}

At the start of the user study, each user completed a practice round with four simple tasks to familiarize them with our experimental setup. Initially, they manually controlled the robot using Teleop to grasp an object. This trial was repeated with SAG to expose users to autonomous assistance. The third task introduced intentional errors in SAG's privileged knowledge to encourage users to correct the robot when needed. The final task demonstrated the robot's efficiency in guiding users past incorrect objects to the desired object, building users' trust in the robot's assistance.
After the familiarization stage, each user completed the following three tasks. A snapshot of the experiment setup of each task is shown in  Figure \ref{fig:experiment-setup}.

\subsubsection{Pick and Place} Users were asked to perform a pick-and-place operation involving two objects within the environment. They were instructed to sequentially place each object on separate designated pedestals. For this task, SAG was provided with perfect knowledge of object locations. The aim of this task was to test if VOSA does better than Teleop and if VOSA does as well as SAG when SAG has this privileged information about user intent.

\subsubsection{Deceptive Grasping} 
This task required users to grasp a target object placed behind two co-planar objects. Here, the oracle knowledge provided to SAG contained only the co-planar object locations and information about the target object was missing. As a result of this misrepresentation, SAG can only infer intent over the co-planar objects, failing to consider the target goal.
This highlights the difficulty of real-world interactions where all the object locations may not be known in advance. This task was designed to be especially easy for Teleop but complicated for SAG. Prior studies have proposed similar ``deceptive" tasks that intentionally require the human to fight for control over a robot that is executing an undesired motion~\cite{aronson2024intentional}. We included this task to evaluate if VOSA, which does not suffer from the same inference issues as SAG, overcomes the issues observed in prior work.

\subsubsection{Shelving} This task seeks to mimic scenarios encountered in real-world settings. 
The users were asked to use the robot to help their roommate store groceries by placing bottled drinks on a designated shelf. During the task, a human places groceries on a table, and the robot must help store all drinks. Over the course of the task, the robot works together in real-time with the human as more groceries are unloaded on the table. This task aimed to introduce dynamic objects into the environment, mimicking real-life situations where the object locations may vary.

\subsection{Study Protocol}
We recruited 18 subjects (14 males, 4 females, average age $25\pm2.99$) from the university campus to participate in an IRB approved study using a within-subjects experimental design. Users provided informed consent prior to their participation. The average duration of the study was 48 minutes per subject, with a compensation of \$25.

\subsection{Metrics Evaluated}
\textit{Objective Metrics:} We recorded the time each user took to complete individual treatments and the magnitude of joystick input exerted during each treatment. This allowed us to assess the speed with which tasks were completed and the level of interaction between the users and the robotic arm.

\textit{Subjective Metrics:} After each treatment, participants were asked to provide responses to the following 7-point Likert scale survey:
(1) How confident are you that the robot \textbf{recognized the objective} of the task?
    (2) How \textbf{in control} did you feel during the task?
    (3) How \textbf{quickly} was the robot able to accomplish the task?
    (4) How \textbf{helpful} was the robot’s behavior?
    (5) How \textbf{trustworthy} was the robot’s behavior?
    (6) How \textbf{frustrating} was this task for you?
    (7) How interested are you in \textbf{collaborating} with the robot again?

After completing all three treatments for a task, the user was also asked to indicate a preference between Treatment I, II, III, or neutral/no preference. Users were blind to the treatment type, and the order of treatments is randomized.

\section{Results}
\begin{table*}[h]
\centering
\caption{Likert Scale Questionnaire Responses from 18 Users (Mean $\pm$ Std. Error) *significant result ($p < 0.05$)}
\label{tab:qualitative-responses}
\begin{tabular}{ccccccccc}
\hline
 &
  Method &
  Recognition ($\uparrow$) &
  Control ($\uparrow$) &
  Speed ($\uparrow$) &
  Help ($\uparrow$) &
  Trust ($\uparrow$) &
  Frustration ($\downarrow$) &
  Future Collab. ($\uparrow$) \\ \hline
\multirow{3}{*}{\shortstack{Deceptive\\Grasping}} &
  Teleop &
  2.44 $\pm$ 0.46 &
  \textbf{5.78 $\pm$ 0.53} &
  4.61 $\pm$ 0.32 &
  2.06 $\pm$ 0.37 &
  4.33 $\pm$ 0.40 &
  \textbf{1.61 $\pm$ 0.27} &
  \textbf{5.72 $\pm$ 0.35} \\
 &
  SAG &
  2.05 $\pm$ 0.25 &
  4.06 $\pm$ 0.50 &
  2.06 $\pm$ 0.37* &
  1.83 $\pm$ 0.28 &
  2.06 $\pm$ 0.33* &
  5.28 $\pm$ 0.42* &
  3.33 $\pm$ 0.59* \\
 &
  VOSA (ours) &
  \textbf{5.05 $\pm$ 0.38*} &
  4.67 $\pm$ 0.36 &
  \textbf{5.17 $\pm$ 0.39} &
  \textbf{5.16 $\pm$ 0.37*} &
  \textbf{5.06 $\pm$ 0.35} &
  2.11 $\pm$ 0.29 &
  5.67 $\pm$ 0.32 \\ \hline
\multirow{3}{*}{Shelving} &
  Teleop &
  1.83 $\pm$ 0.38* &
  \textbf{6.22 $\pm$ 0.45} &
  4.56 $\pm$ 0.47 &
  2.17 $\pm$ 0.44* &
  4.17 $\pm$ 0.37 &
  2.5 $\pm$ 0.44 &
  5.17 $\pm$ 0.44 \\
 &
  SAG &
  5.05 $\pm$ 0.31* &
  4.50 $\pm$ 0.32 &
  5.11 $\pm$ 0.38 &
  5.0 $\pm$ 0.23* &
  4.77 $\pm$ 0.27 &
  2.56 $\pm$ 0.35 &
  5.44 $\pm$ 0.33 \\
 &
  VOSA (ours) &
  \textbf{5.88 $\pm$ 0.28*} &
  4.61 $\pm$ 0.47 &
  \textbf{6.0 $\pm$ 0.28*} &
  \textbf{6.11 $\pm$ 0.24*} &
  \textbf{5.5 $\pm$ 0.22*} &
  \textbf{2.06 $\pm$ 0.37} &
  \textbf{6.22 $\pm$ 0.25} \\ \hline
\end{tabular}
\end{table*}

\begin{figure}
    \centering
    \includegraphics[width=.7\linewidth]{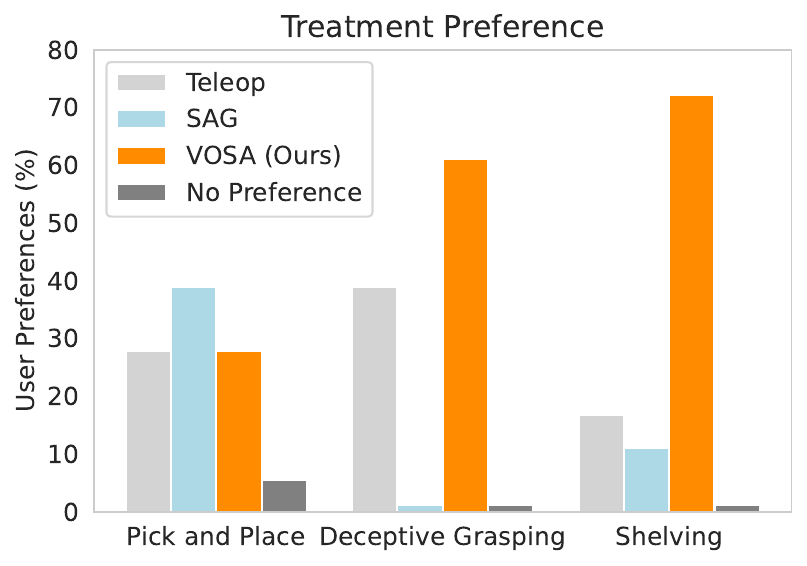}
    \caption{\textbf{Assistance Preferences} for three manipulation tasks. For deceptive grasping and  shelving tasks, a majority of users prefer VOSA (our method) over a shared autonomy baseline (SAG) and direct teleoperation (Teleop).}
    \label{fig:preferences}
\end{figure}

%results
The results from the user study are shown in Figures~\ref{fig:quant-results}-~\ref{fig:preferences} and Table~\ref{tab:qualitative-responses}. The user questionnaire showed no significant results for the pick and place task. Therefore, pick and place results were omitted from Table \ref{tab:qualitative-responses}.
We analyze our hypotheses using a single-factor repeated measures ANOVA to determine a difference in outcomes, using the treatment (assistance method) as the independent variable. For comparisons where ANOVA rejects the null hypothesis, we perform post-hoc pairwise comparisons across treatments using paired two-tailed t-testing to evaluate significance. We use a significance level of 0.05 for all comparisons. In cases where there is not a significant difference in outcomes, we demonstrate that one treatment is not inferior to another using non-inferiority testing \cite{lesaffre2008superiority}. In each non-inferiority test, we provide a percentage of the difference between treatment means and analyze the difference on a per-task basis.

% \begin{table}[]
% \centering
% \caption{\edit{Task Success Rates}}
% \label{tab:success-rates}
% \begin{tabular}{|l|ccc|}
% \hline
%                    & \multicolumn{3}{c|}{Teleoperation Method}  \\
% Task               & Teleop       & SAG          & VOSA (ours)         \\ \hline
% Pick \& Place       & \textbf{1.0} & 0.88         & 0.94         \\
% Deceptive Grasping & \textbf{1.0} & 0.94         & \textbf{1.0} \\
% Shelving           & 0.88         & \textbf{1.0} & \textbf{1.0} \\ \hline
% \end{tabular}
% \end{table}

Throughout the study, a total of 6 task failures occurred. We defined task failure as participants knocking over objects or pushing the target intent outside of the manipulator's workspace. 
Three failures were recorded on the pick and place task: one user failed using VOSA and two users failed using SAG. Two users failed the Shelving task using Teleop and one user failed the deceptive grasping task using SAG. To ensure fair within-subjects statistical tests, participants were allowed to re-attempt the task in cases of failure.

\subsection{Hypothesis Testing}
H1 and H2 test the performance of VOSA when compared to Teleop and SAG in the pick and place task. Recall that this task provides the SAG baseline perfect oracle access to all possible user intents, making SAG a near-optimal assistance method. For task completion time (H1), ANOVA revealed that there is a significant difference between treatments ($F(2, 51)=5.14, p=.009$) and post-hoc testing verified that VOSA significantly accomplishes this task in less time than the teleop method ($p=0.03$). When comparing VOSA to SAG (oracle), a non-inferiority test shows that VOSA lies within a 10.5\% (time=8.27s) non-inferiority margin of SAG, which was provided information about all possible intents prior to task execution. Therefore, our results demonstrate that VOSA can assist humans in completing tasks faster than Teleop and is not significantly slower than SAG, supporting H1.

% In our study, the treatment order was randomized, but the task order was always consistent, sorted from easiest to hardest, starting with Pick and Place. We believe that the large variance within each treatment for Pick and Place is due to users still familiarizing themselves with our robot.

% As a result, our metrics experienced large variance for this task, preventing any statistical acceptance of H1.

For H2, we find a significant difference in the amount of human input required to complete the pick and place task ($F(2, 51)=30.19, p<.001$)  (Figure \ref{fig:total-effort}). In this task, SAG and VOSA required significantly lower input magnitudes when compared to Teleop ($p < .001$). We also find that VOSA lies within a 16.3\% (effort=36.0) non-inferiority margin of SAG. Therefore, there is weak support for H2. We predict that the slight increase in input magnitude for VOSA compared to SAG comes from the need to offer slight corrections to VOSA more frequently than SAG. Recall that SAG has perfect knowledge of all possible human intent, which results in much smoother trajectories than those generated under VOSA, which predicts (with some noise) the human's intent repeatedly at runtime.
 
H3 considers the use of assistance in situations where the ground truth intent is underrepresented in the oracle information provided to SAG. We studied this in two ways: \textit{intent inaccuracy} and \textit{intent incompleteness}. 

\textit{Intent inaccuracy} occurs in the shelving task, where the intent is correct at the start of the task, but then objects in the scene are perturbed, and the original intent information is no longer correct. In the case of inaccurate intent knowledge, we find that there is a significant difference between treatments for both time ($F(2, 51)=17.47, p<.001$) and effort ($F(2, 51)=106.36, p<.001$). For task duration, post-hoc testing shows that compared to Teleop, VOSA and SAG were significantly faster (both $p<.001$). For human effort, the baseline and VOSA each resulted in significantly lower input magnitudes compared to Teleop ($p < .001$). Furthermore, non-inferiority measures indicate that VOSA requires 1.7\% less time and 7.7\% less input than SAG. Together, this indicates strong non-inferiority for VOSA over the oracle method in cases of intent inaccuracy. However, VOSA does not statistically outperform SAG for this task. Therefore, the data obtained in the Shelving task does not support H3 directly. However, it should be noted that in real-world scenarios of zero-shot inference, the baseline shared autonomy would not know the set of user intents and, therefore, would perform no better than Teleop. 

The second form of misrepresentation, \textit{intent incompleteness}, occurs in the deceptive grasping task. Here, the user's true intent is missing entirely from the oracle knowledge provided to SAG. In this task, ANOVA reveals a significant difference in the measured time ($F(2, 51)=39.9, p<.001$) and input magnitude ($F(2, 51)=56.7, p<.001$) for the three treatments on this task. Pairwise testing reveals that VOSA significantly outperforms SAG in both time and human effort (both $p<.001$). Therefore, the data obtained from this task confirms H3 in cases of intent incompleteness. 

For H4 and H5, we consider the qualitative assessment of manipulation assistance based on user survey responses (Table~\ref{tab:qualitative-responses}).
In the shelving task, users rate VOSA more highly than SAG for both intent recognition ($p=.003$) and helpfulness ($p<.001$). Similarly, in deceptive grasping, users reported that VOSA continued to provide better intent recognition ($p<.001$) and helpful assistance ($p<.001$) when compared to the feedback that SAG received.
These findings establish strong support for both H4 and H5.

\subsection{Study Insights}
\textit{1) VOSA is the preferred method of assistance for tasks with intent misspecification and semi-autonomous assistance is generally preferred overall.}
Figure \ref{fig:preferences} shows the users' preferred treatment for each task.  Teleop and VOSA were equally preferred for the pick and place task, while SAG was slightly more favored, and one user had no preference, suggesting that for simple tasks, users find traditional control methods satisfactory and slightly prefer oracle methods that have perfect knowledge. VOSA is generally preferred in shelving and deceptive grasping, where intents are dynamic in nature.
Notably, in deceptive grasping, Teleop is still preferred by 38\% of users, and no user indicated a preference for SAG. No reported preferences for SAG in this task indicates that either A) assistance from a robot that cannot infer user intent correctly is generally undesirable or B) many humans prefer to stay in control of the robot during difficult tasks. We elaborate on the former in Insight 2 and the latter in Insight 3.

\textit{2) Incomplete intent specification results in catastrophic SAG performance.}
The deceptive grasping task showcases the pitfalls that occur with many prior shared autonomy approaches when the user tries to accomplish an unknown intent~\cite{zurek2021situational, aronson2024intentional}---SAG significantly hinders the user to the point that Teleop alone performs significantly better than it. For the deceptive grasping task, the target intent was placed behind two deceptive objects, which were the only intents provided to SAG before the task (Figure \ref{Deceptive reaching}). Here, SAG does not have a reason to help the user direct the robot beyond the first two objects. The target intent is placed in close proximity to, and between, the two known objects, so the robot always assigns high confidence to an incorrect intent. It is only because we never give complete control ($\alpha \le 0.8$, Fig. \ref{fig:Arbitration}) to the robot that the user can fight for control of the confused robot with success. One user even believed that this task was impossible using SAG, stating, ``The robot is stuck in an infinite loop!'' referring to the robot alternating commands between the two known intents while the user was desperately attempting to command the robot toward the correct intent. 

Prior work has addressed the problem of missing/unknown intents though corrective demonstration feedback~\cite{zurek2021situational}, or by asking users to fight against the robot as it moves to incorrect intents~\cite{aronson2024intentional}. By contrast, the power of the VOSA framework is that it has the potential to enable teleoperation assistance even if the user is targeting an intent that has never been seen before. Deceptive grasping highlights several significant outcomes, indicating that users are very confident in VOSA when it comes to recognition and help and very disapproving of SAG when it comes to speed, trust, and desire to collaborate.

\textit{3) Some users prefer no assistance.}
% Users who preferred Teleop for more difficult tasks, such as shelving, consistently favored Teleop across all task types. 
An interesting finding from our study was that out of the 18 users, 3 users indicated a preference for Teleop in every task. This is particularly surprising, given the simple tabletop manipulation tasks we defined and the fact that the first task has a near-optimal form of SAG assistance. Interestingly, these users also exhibited higher-than-average input magnitudes across all tasks, even though they were taught in training to trust the robot when they felt like it was executing the task correctly. On average, these users rated Teleop highest in terms of trust, 6.0 on a scale of 7-point Likert scale, while providing lower trust ratings for SAG and VOSA, 3.8 and 4.0 respectively, across all the tasks. These findings suggest a strong preference for maintaining control, aligning with trends observed in previous literature~\cite{javdani_shared_2015, Gombolay-RSS-14, 5941028}. Future studies should explore human-robot trust in context of the VOSA framework and evaluate how to augment shared autonomy methods to foster greater trust in the robot.
%%%%%%%%%%%%%%%%%%%%%%%%%%%%%%%%%%%%%%%%%%%%%%%%%%%%%%%%%%
% User analysis
% People who preffered Teleop in long horizon task also preffered teleop in all the 3 tasks.
% Trust - Overall trend, Trust in VOSA more when users subjected to Teleop or Baseline first.
% No particular insights on ordering bias. Some users clocked high input magnitude for VOSA when subjected to baselines first, but others trusted VOSA and clocked significantly lower input magnitude.
% Irrespective of order and task, users found VOSA to be more helpful in Task 2 and Task 3 than baselines and equally helpful (wrt SAG) in Task 0).
% With increased experience users became more comforatble with the robot.
% Task 1 people who preffered teleop had <= average(+-50) input magnitude of Task 1. Same for Task 2 for the same set of people. same for Task 3.
% All the people who preffered teleop in atleast one of the task had higher input magnitude than the average in all the tasks.
% For Task 1 most of the people had >= avergae trust score. Also true for Task 2 (SAG and VOSA). Not True for Task 3, average trust score was 5 and most of the users rated 5 or 6 for VOSA. But Largely same story for SAG.
%%%%%%%%%%%%%%%%%%%%%%%%%%%%%%%%%%%%%%%%%%%%%%%%%%%%%%%%%%
\section{Conclusion and Future Work}
We introduced Vision-Only Shared Autonomy (VOSA), a simple yet effective approach to achieve zero-shot shared autonomy in environments with all or partially unspecified human manipulation intents. VOSA has three main components: leveraging point cloud data from only a single end-effector-mounted camera, real-time intent inference, and arbitrating between human and user inputs.

% would be three main sections of VOSA, which this combination makes VOSA surpass traditional methods that rely on predefined all possible intets.
% Leveraging RGBD point cloud data from an end-effector-mounted camera, VOSA enables real-time intent inference, surpassing traditional methods that rely on predefining all possible intents.
Through extensive evaluation in a user study involving 18 participants across three tasks, we demonstrated that VOSA not only closely matches the performance of traditional shared autonomy systems in known intent scenarios but also outperforms them in settings where intents are not fully known \textit{a priori}. Our work takes a step towards achieving zero-shot, high-performance, adaptive shared autonomy, offering a scalable and efficient solution for a wide range of assistive robotics applications. Our findings underscore the importance of integrating vision-based techniques with shared control paradigms, laying the groundwork for future advancements in a human-robot collaboration that can adapt to user intentions and environmental contexts in real-time. 

We demonstrated a simple, but effective, version of zero-shot shared autonomy, but its potential could be expanded with the following improvements to our three components:
 
\textbf{Perception:} Our current perception system faces practical challenges common to computer vision, such as sensitivity to lighting conditions and occasional difficulties with object detection using YOLO. Further integration of advanced computer vision techniques could enhance the system's understanding of the environment. For example, utilizing object detection models and depth-image template matching ~\cite{muelling2015autonomy} and leveraging perception systems that incorporate a dynamic environment representation ~\cite{reddy_shared_2018, esfandiari2024bimanualmanipulationsteadyhand}, would improve the robustness of object detection and facilitate real-time adjustment to the robot's actions based on the evolving scene. 
% Additionally, vision-language models~\cite{ zhang2024vision} can enhance prediction and scene understanding while maintaining zero-shot capabilities.
Additionally, vision-language models~\cite{ zhang2024vision} can be utilized to enhance prediction and scene understanding while still upholding the zero-shot nature of VOSA.
% the understanding of objects in the environment and their placement intents. 
% These models have the potential to provide insights into object geometries and contextual cues to better facilitate shared autonomy 
%while still upholding the zero-shot nature of VOSA.
 
\textbf{Prediction:} Incorporating and developing more sophisticated intent inference methods and learning-based approaches would improve the robot's ability to anticipate user needs in dynamic environments. Future work could further refine the prediction by incorporating models of human intent learned from demonstrations and leveraging historical data to enhance the system's predictive accuracy ~\cite{reddy_shared_2018,jonnavittula_sari_2023,zurek2021situational,mehta2022learning,yoneda_noise_2023}. Furthermore, using natural language for intent disambiguation could make human input more intuitive and enable better understanding and prediction for user needs in real-time ~\cite{nikolaidis2018planning,mo2023towards,yow_shared_2024}; however, natural language is not a viable interface for many disabled users.
  
\textbf{Arbitration:} Adopting a blending approach where control is shared dynamically between the user and system based on the system's confidence level allows for smooth, adaptive transitions. This flexible arbitration can be achieved by defining different arbitration functions based on task complexity and user preference, ensuring that the system adapts to both novice and expert user's need~\cite{muelling2017autonomy, dragan2013policy, downey2016blending, gopinath2016human}. Also, there is potential to learn arbitration strategies using user interaction data, improving adaptability by refining blending through hindsight data aggregation~\cite{oh2019learningarbitrationsharedautonomy} or using deep reinforcement learning~\cite{9727522}.

\section{Acknowledgment}
This work has taken place in the Aligned, Robust, and Interactive Autonomy (ARIA) Lab at the University of Utah. ARIA Lab research is supported in part by the NSF (IIS-2310759, IIS2416761), the NIH (R21EB035378), ARPA-H, Open Philanthropy, and the ARL STRONG program.

\bibliographystyle{unsrt}
\balance
\bibliography{main}

% \bibitem{b1} G. Eason, B. Noble, and I. N. Sneddon, ``On certain integrals of Lipschitz-Hankel type involving products of Bessel functions,'' Phil. Trans. Roy. Soc. London, vol. A247, pp. 529--551, April 1955.
% \bibitem{b2} J. Clerk Maxwell, A Treatise on Electricity and Magnetism, 3rd ed., vol. 2. Oxford: Clarendon, 1892, pp.68--73.
% \bibitem{b3} I. S. Jacobs and C. P. Bean, ``Fine particles, thin films and exchange anisotropy,'' in Magnetism, vol. III, G. T. Rado and H. Suhl, Eds. New York: Academic, 1963, pp. 271--350.
% \bibitem{b4} K. Elissa, ``Title of paper if known,'' unpublished.
% \bibitem{b5} R. Nicole, ``Title of paper with only first word capitalized,'' J. Name Stand. Abbrev., in press.
% \bibitem{b6} Y. Yorozu, M. Hirano, K. Oka, and Y. Tagawa, ``Electron spectroscopy studies on magneto-optical media and plastic substrate interface,'' IEEE Transl. J. Magn. Japan, vol. 2, pp. 740--741, August 1987 [Digests 9th Annual Conf. Magnetics Japan, p. 301, 1982].
% \bibitem{b7} M. Young, The Technical Writer's Handbook. Mill Valley, CA: University Science, 1989.
% \end{thebibliography}
% \vspace{12pt}
% \color{red}
% IEEE conference templates contain guidance text for composing and formatting conference papers. Please ensure that all template text is removed from your conference paper prior to submission to the conference. Failure to remove the template text from your paper may result in your paper not being published.

\end{document}